\documentclass{article}

\usepackage{arxiv}
\usepackage[utf8]{inputenc}                         
\usepackage[T1]{fontenc}                            
\usepackage{hyperref}                               
\usepackage{url}                                    
\usepackage{booktabs}                               
\usepackage{amsfonts}                               
\usepackage{nicefrac}                               
\usepackage{microtype}                              
\usepackage{graphicx}
\usepackage[labelfont=bf]{caption}                  
\usepackage[superscript, biblabel]{cite}            
\usepackage{float}                                  
\usepackage{xcolor}
\usepackage{placeins}                               

\title{Improving the Performance of Fine-Grain Image Classifiers via Generative Data Augmentation}

\date{April 27, 2020}	

\author{ 
    {Shashank Manjunath} \\
	Charles River Analytics\\
	Cambridge, MA, 02138\\
	\texttt{smanjunath@cra.com} \\
    \And
    {Aitzaz Nathaniel} \\
    U.S. Army CCDC C5ISR Center \\
    Aberdeen Proving Ground, MD, 21005 \\
    \texttt{aitzaz.nathaniel.civ@mail.mil} \\
	\And
	{Jeff Druce} \\
	Charles River Analytics\\
	Cambridge, MA, 02138\\
	\texttt{jdruce@cra.com} \\
    \And
	{Stan German} \\
	Charles River Analytics\\
	Cambridge, MA, 02138\\
	\texttt{sgerman@cra.com} \\
}

\renewcommand{\headeright}{}
\renewcommand{\undertitle}{Technical Report}

\begin{document}
\maketitle

\begin{abstract}
Recent advances in machine learning (ML) and computer vision tools have enabled applications in a wide
variety of arenas such as financial analytics, medical diagnostics, and even within the Department of Defense. However,
their widespread implementation in real-world use cases poses several challenges: (1) many applications are highly
specialized, and hence operate in a \emph{sparse data} domain; (2) ML tools are sensitive to their training sets and
typically require cumbersome, labor-intensive data collection and data labelling processes; and (3) ML tools can be
extremely “black box,” offering users little to no insight into the decision-making process or how new data might affect
prediction performance. To address these challenges, we have designed and developed Data Augmentation from Proficient
Pre-Training of Robust Generative Adversarial Networks (DAPPER GAN), an ML analytics support tool that automatically
generates novel views of training images in order to improve downstream classifier performance. DAPPER GAN leverages
high-fidelity embeddings generated by a StyleGAN2 model (trained on the LSUN cars dataset) to create novel imagery for
previously unseen classes. We experimentally evaluate this technique on the Stanford Cars dataset, demonstrating
improved vehicle make and model classification accuracy and reduced requirements for real data using our GAN based data
augmentation framework. The method’s validity was supported through an analysis of classifier performance on both
augmented and non-augmented datasets, achieving comparable or better accuracy with up to 30\% less real data across
visually similar classes. To support this method, we developed a novel augmentation method that can manipulate
semantically meaningful dimensions (e.g., orientation) of the target object in the embedding space.
\end{abstract}


\section{Introduction} \label{sec:intro}
Machine learning (ML) based image classifiers have seen widespread adoption across multiple industries, such as financial
analytics, medical diagnostics, and defense within the last few years. However, these tools are frequently trained on
insufficient data for their particular application, leading to poor performance in production environments.
Cutting-edge ML algorithms have recently emerged to assist with image analysis. These capabilities
support analysis by automatically and accurately performing inference on data (e.g., automatic target recognition (ATR),
target tracking, image segmentation). High-volume, hand-labeled datasets are typically required for satisfactory
training and testing to benefit from the capabilities of ML. However, in many real-world scenarios, ML algorithms fail to
perform due to insufficient data volume and/or diversity. For example, consider the case of a binary classifier trained
to recognize a specific vehicle make and model using images of a single instance of this vehicle. Regardless of the number of images of this one vehicle, the classifier
will likely struggle to generalize its classification capability to identical makes and models of different colors, from novel perspectives, and backgrounds.

Existing techniques to train and validate high-performing ML tools do not extend to scenarios where only sparse
datasets, or datasets with insufficient data diversity, are available. Data augmentation is a common approach for
artificially increasing the volume and diversity of available imagery, thereby improving task performance (e.g., image
classification). Traditional data augmentation approaches expand sparse datasets in an unsupervised manner; for example,
by applying a series of affine transformations (e.g., rotations and/or shear) or additive noise to a sparse set of images. However, these
methods typically achieve modest performance gains where a reasonable amount of data is already
available.~\cite{wong_understanding_2016} These approaches can be effective, but since they are unsupervised, no
explicit guiding force promotes the production of new data to enhance the ML tool performance. A more principled
approach will explicitly take the ultimate data augmentation goal (better performance for ML tools) into consideration.

The main goal of this work is to design an algorithm which generates novel, semantically meaningful data
from sparse, homogeneous (insufficiently diverse) datasets to enhance ML algorithm performance. Our contributions
are summarized as follows:
\begin{enumerate}
  \item We develop a StyleGAN2-based~\cite{karras_analyzing_2019} technique to generate augmented data with novel characteristics.
  \item We show that the given augmented improves classifier performance on fine-grain image
    classification tasks using the Stanford Cars dataset~\cite{krause_3d_2013}.
  \item We verify that the augmented data enhances classifier decisions and focuses classifiers on discriminative
    features through Grad-CAM attention maps~\cite{selvaraju_grad-cam_2020}.
\end{enumerate}

\section{Related Work} \label{sec:related_work}
Computer vision tools have demonstrated human level performance for various applications and are
routinely deployed successfully in the wild (e.g., face identification, object classification, image
segmentation)~\cite{schmidhuber_deep_2015}. A primary source for this impressive performance is the resurgence of Deep Neural Networks (DNNs),
which employ extremely large convolutional neural networks (CNNS) to achieve state-of-the-art
results~\cite{lecun_deep_2015}. The volume of the datasets used to train these powerful models has scaled up to
meet the requirements of the ever larger modeling capacities of these networks (e.g., some classes in ImageNet contain
over 10K unique images)~\cite{russakovsky_imagenet_2015}. Straightforward affine transformation (e.g., scaling and
shearing) of training samples has become standard practice for augmenting training sets, but are not sufficient to overcome
significant data sparsity. 

Another approach leverages generative models to bolster training sets and ultimately improve
performance~\cite{perez_effectiveness_2017}. GANs are generative models trained by pitting two networks against each other: the first learns to synthesize novel “counterfeit” images, which attempt to fool a second discriminator network that learns to identify counterfeit samples~\cite{goodfellow_generative_2014}. GANs have benefited
from a surge in popularity across a variety of applications, in part due to architecture refinements enabling the production
of photorealistic images of usable size~\cite{radford_unsupervised_2016-1}, and the ability to add class attributes to
input via conditional GANs~\cite{mirza_conditional_2014}. Improvements in training procedures (training is notoriously
difficult in GANs), such as loss function enhancements, have also led to increased
performance. Conditional GANs also emerged and functioned by including labels into the model input, sculpting output to
match (e.g., text to image generation, where the text dictated the style of the output). For data augmentation,
transferring GAN models from comparable domains has recently been investigated~\cite{antoniou_data_2018}. 

To improve the performance of fine-grained machine leaning classifiers using traditional supervised learning techniques,
we require augmentation of specific, fine-grained classes. Existing GAN techniques aim to generate high-resolution imagery
of specific classes. Conditional GANS (cGANS) aim to precondition generated data with specific
characteristics~\cite{mirza_conditional_2014}. cGANS have demonstrated state-of-the-art performance on many
generative techniques, have previously been used for data augmentations, and can learn meaningful latent
embeddings from training data. However, these latent embeddings are frequently highly
entangled\cite{shen_interpreting_2020}. Therefore, generating imagery of previously unseen classes is exceedingly difficult. Our
use case requires conditioning the generation of specific, fine-grain details from classes which either have
not been seen by the generative model, or have only been seen in a small quantity. cGANS, as with most GAN-type models,
are known to require large amounts of accurately labelled data for training\cite{goodfellow_generative_2014}.
Furthermore, these models do not allow generation of specific, previously unseen characteristics from our dataset, such
as novel paint colors on a Toyota Camry. In this paper, we use StyleGAN2 to generate novel views of previously unseen
image data for use in training fine-grain image classifiers.

\section{Methods} \label{sec:methods}
The  DAPPER GAN system takes in a novel, unseen example of an object, creates a latent space embedding based on
this image, and generates a modified image of the object to augment the training set and increase classification performance.

\subsection{Embedding Creation} \label{subsec:embedding_creation}
\begin{figure}[!hbt]
    \centering
    \includegraphics[width=.5\textwidth]{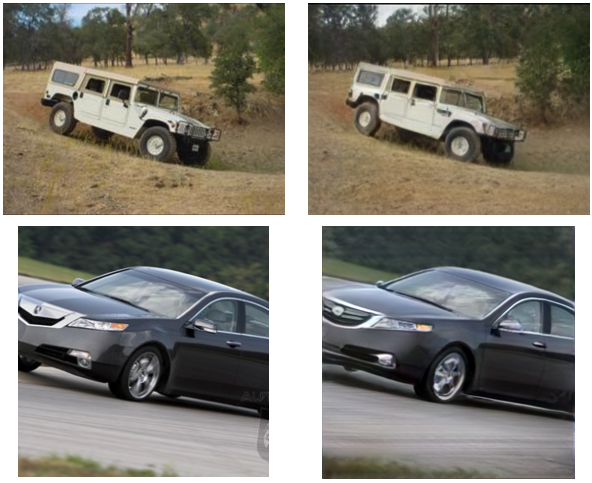}
    \caption{\textbf{Example StyleGAN2 reconstructed vehicle images.} \emph{(Left column) Original images from the
    Stanford Cars dataset. (Right column) Reconstructed images by StyleGAN2 from backprojected latent vectors}}
    \label{fig:stylegan_reconst}
\end{figure}

\begin{figure}[!hbt]
    \centering
    \includegraphics[width=.6\textwidth]{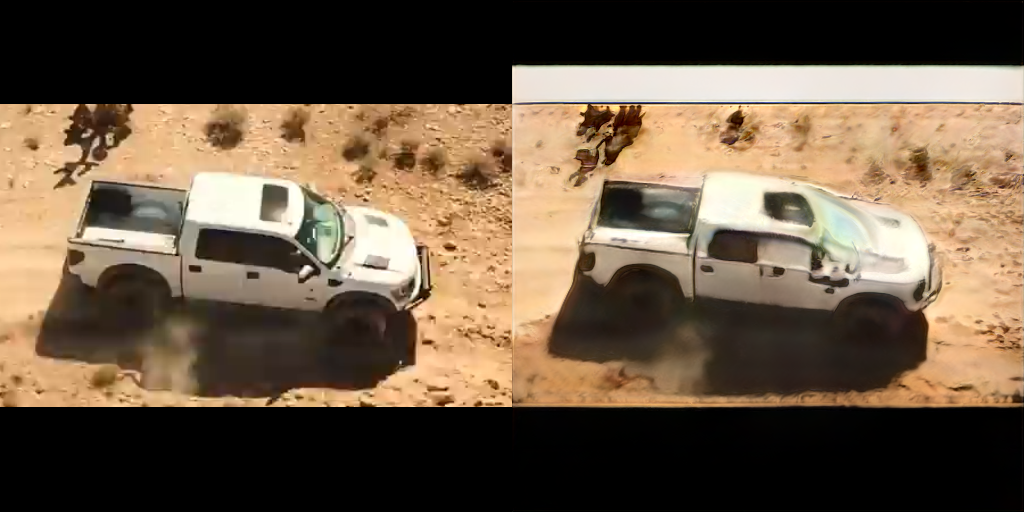}
    \caption{\textbf{Example reconstructions of real, out-of-distribution aerial data.} \emph{(Left) Original image.
    (Right) Reconstructed Image. By searching the latent space of StyleGAN2 using an optimization technique, as
    described by Karras et. al., we can find latent variables that enable generation of arbitrary vehicles of interest,
    even if the vehicle is not represented in the training dataset.~\cite{karras_analyzing_2019}}}
    \label{fig:sponsor_reconst}
\end{figure}

Transfer from one dataset to another is a common problem in machine learning which many researchers have attempted to
address. Using DAPPER GAN, we have shown the ability to reuse learned representations across disparate datasets, e.g.,
applying a StyleGAN2 model trained on the LSUN Cars dataset~\cite{yu_lsun_2016} to the Stanford Cars
dataset.~\cite{krause_3d_2013} To accomplish this task, we implemented a projection technique that searches StyleGAN2’s
low-dimensional intermediate latent space, $W$, for latent representations which most faithfully reconstruct the original
high-dimensional image~\cite{karras_analyzing_2019} By freezing the weights of the generator, incorporating a loss
between the image of interest and the generated image, and using an optimization function to update the latent
representation to decrease the loss, we can obtain an optimal latent vector and exceptionally accurate reconstructions
from sparse input imagery. Example reconstructions from overhead data are shown in Figure \ref{fig:sponsor_reconst}, and
reconstructions of data from the Stanford Cars dataset are shown in Figure \ref{fig:stylegan_reconst}. Recall that the
StyleGAN2 generator was not trained on these images, but rather on the LSUN Cars dataset, which is a large, unlabeled
dataset of vehicles that explicitly excludes classes, such as pickup truck. Furthermore, the generator was not trained
on overhead data, but still is able to recreate overhead images of the given pickup truck.

\subsection{Data Generation}
\label{subsec:data_generation}
We used a StyleGAN2 based generator~\cite{karras_analyzing_2019}, which is capable of generating high-resolution
(1024x1024) realistic imagery. As an unsupervised technique, StyleGAN2 learns from unlabeled imagery. This feature is of
significant benefit to DAPPER GAN as no data labelling is required, only a large corpus of reasonably representative
images, called the \emph{source dataset}. Furthermore, the source dataset does not need to contain the particular class
of interest, but rather generally similar classes. For example, although the generator was not trained on images of
pickup trucks (in fact the LSUN dataset specifically excludes pickup trucks), we are still able to find latent vectors
which can realistically generate pickup trucks~\cite{yu_lsun_2016}. Furthermore, although the LSUN dataset does not
contain aerial data, the system can generate data from an aerial view. We successfully used a StyleGAN2-based data
augmentation technique to generate realistic data of vehicles of interest with high enough fidelity to identify specific
vehicle makes and models, as showin in Figure \ref{fig:stylegan_reconst} and Figure \ref{fig:sponsor_reconst}.
StyleGAN2 allows both generation of highly realistic images and projection of images into an intermediate latent space.
The former attribute is useful for generating images with discriminative features, while the latter feature allows us to
generate images of a known class.  

\begin{figure}[!hbt]
    \centering
    \includegraphics[width=.65\textwidth]{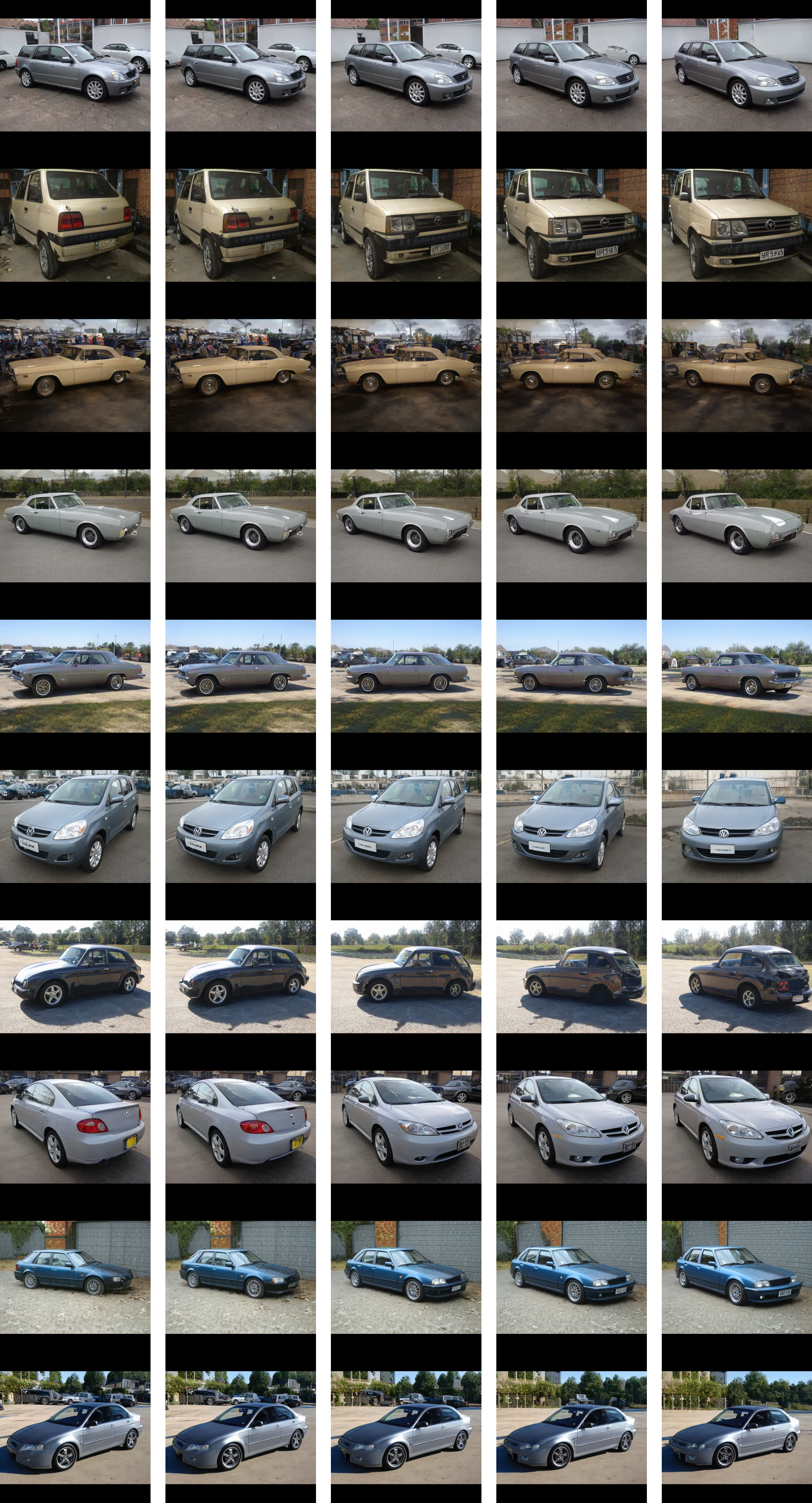}
    \caption{\textbf{Example generated images through perturbations in latent space.} \emph{(Columns from left to
            right): Rotation coefficient -30, rotation coefficient -20, rotation coefficient 0, rotation coefficient 20,
            rotation coefficient 30. The images at Coefficient 0 are the original generated images. Changing the
            coefficient allows rotation of the vehicle in the image, leading to novel views of the car. While we can
            make small, controlled rotations of the car in some cases , we occasionally see rapid and
            uncontrollable changes in orientation.}}
    \label{fig:rotations}
\end{figure}

The ability to find latent representations to generate arbitrary images has enormous implications for the DAPPER GAN
system. Typical ML tool training follows a “supervised learning” paradigm and requires both representative data and
corresponding class labels (e.g., an image of an Audi A4 sedan, and a corresponding class label indicating that the
image contains an Audi A4 Sedan). Since we can find valid latent representations for a previously unseen object of
interest, we can also generate novel synthetic data for that object from one or more images by manipulating the object’s
latent space projection. While we can randomly perturb an object’s latent projection (and this does improve classifier
performance as shown in Figure \ref{fig:classifier_perf}), we risk changing the appearance of the object due to the
unsupervised nature of the perturbation. Instead, we have developed an innovative method for finding semantically
meaningful directions within the latent space, which preserve the identity of the object. Using this method, we have
shown we can manipulate vehicles in 3D from a single image (i.e., through rotation), as illustrated in Figure
\ref{fig:rotations}.

\begin{figure}[!hbt]
    \centering
    \includegraphics[width=.9\textwidth]{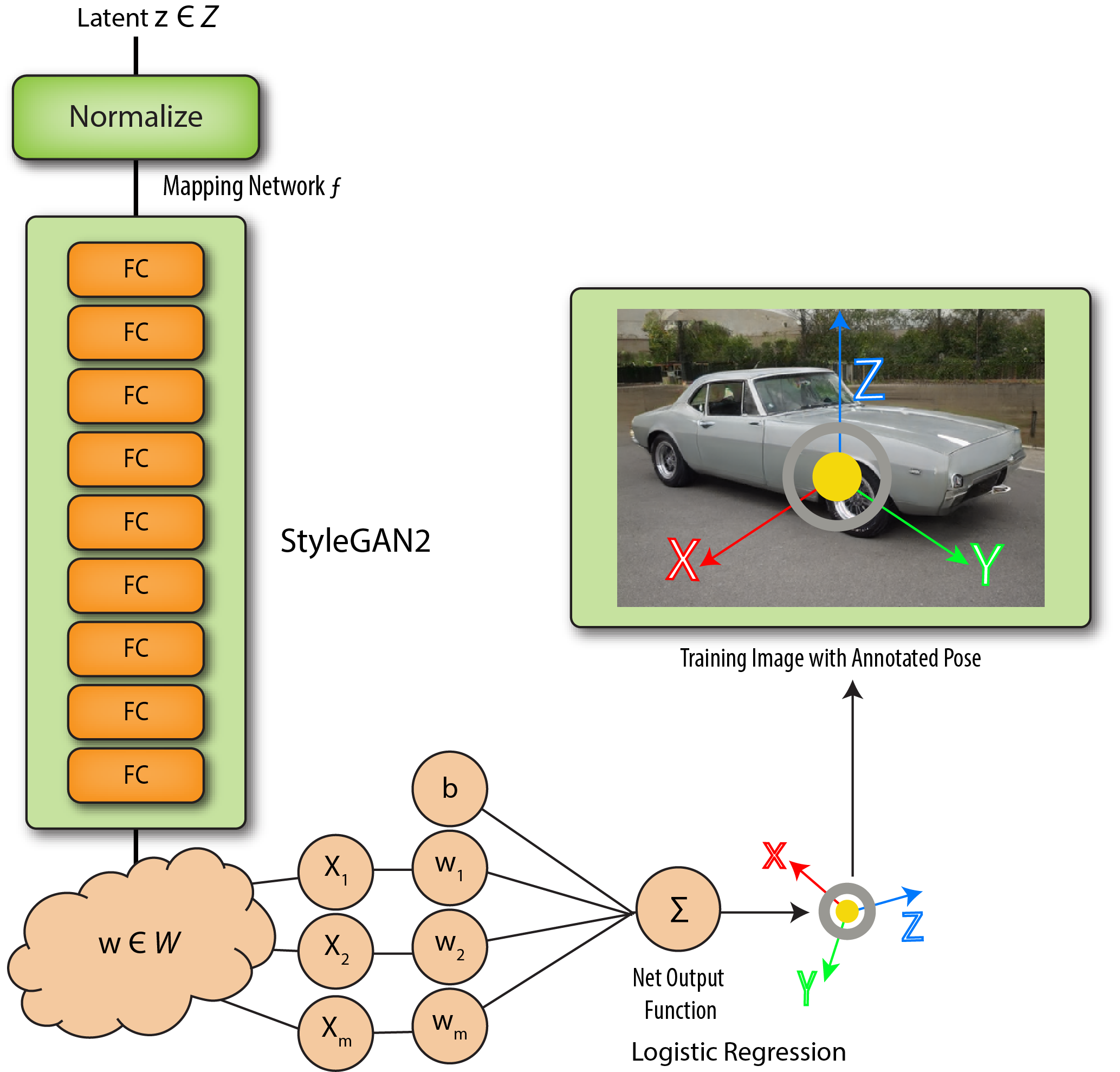}
    \caption{
        \textbf{Illustration of the semantic direction discovery pipeline.}
    }
    \label{fig:rotation_technique}
\end{figure}

\begin{figure}[!hbt]
    \centering
    \includegraphics[width=\textwidth]{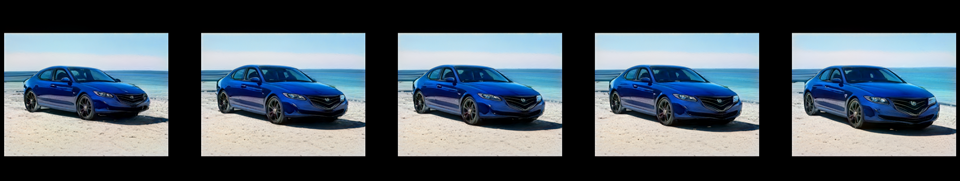}
    \caption{\textbf{Illustration of semantic direction technique applied to data from Stanford Cars.} \emph{In this
    image, we took an image from the Stanford Cars dataset and generated a latent representation for it. We then applied
    our semantic direction discovery technique to the image to generate novel views of the data, which were not seen by
    StyleGAN2 during the network training phase. The reconstruction from latent space without any rotation applied is
    shown in the center image, while novel views are shown in the two leftmost and two rightmost images.}}
\end{figure}

The full rotation technique implements a StyleGAN2 based input projection, which searches the intermediate latent space,
W, of the model for the latent vector that best generates a given data point. This technique allows identification of
latent variables which correspond to particular fine-grained classes (e.g., the latent encoding of a blue 2017 Chevy
Camaro). Furthermore, we have developed a novel technique during our initial effort that enables semantically meaningful
manipulation of StyleGAN2’s $W$ latent space without changing the identity of the target object. We illustrate this
supervised method in Figure \ref{fig:rotation_technique}, which requires large quantities of semantic annotations (e.g.,
object pose), corresponding images and $W$ latent projections to train. To realize this data requirement, we leverage our
StyleGAN2 model to generate 25,000 random synthetic vehicle images, which we filter using an off-the-shelf vehicle
detector. We also train an expert pose estimator using data from the KITTI Vision Benchmark dataset to annotate the
synthetic imagery~\cite{juranek2015real}. We then train another linear predictor to directly estimate the orientation
of each W latent vector within our synthetic dataset. Finally, we threshold and extract the weights from the trained
predictor ($w_1, w_2, … w_m$), which are directly correlated with the scaled contribution of each dimension in $W$ to
its reconstructed vehicle pose, as our latent directional pose vector. To rotate a real vehicle, we simply traverse this
direction vector in the vehicle’s projected $W$ space and reconstruct the image.

\subsection{Explainable Classifiers}
\label{subsec:xai}

While we are able to generate novel imagery of a given imagery and use it to improve classifier performance, we want to
ensure that the generated imagery actually hardens classifier boundaries and leads to classifiers using better
discriminative features when deciding what class imagery belongs to. To this end, we implement gradient-based class
attention mapping (Grad-CAM) to identify specific image features that our classifier uses to decide what class a car
image belongs to~\cite{selvaraju_grad-cam_2020}. Grad-CAM uses the gradients of the final convolutional layer of our
classifier to indicate which regions in an image are important for predicting class, visually showing the areas of the
image that most contribute to classifier decisions.

\FloatBarrier

\section{Results}
\label{sec:results}
The proposed method’s validity was supported through an analysis of classifier performance on both augmented and
non-augmented datasets, achieving comparable or better accuracy with up to 30\% less real data across 10 visibly similar
vehicle models (e.g., Acura TL vs. Acura TSX, BMW X3 vs BMW x5). We additionally demonstrated the ability to transfer
knowledge across datasets, training a StyleGAN2 embedding on the LSUN benchmark dataset and applying it to the Stanford
cars dataset. Furthermore, inspired by recent work from~\cite{voynov_unsupervised_2020}, we developed a novel
augmentation method that can manipulate semantically meaningful dimensions of the target object and its environment
(e.g., object orientation, color, background). This method currently enables DAPPER GAN to generate novel views of the
target object with a single sample image, which we expect will dramatically improve downstream ML tool performance.
Lastly, we showed that augmenting our dataset hardened classifiers and improved explainability of the models, leading to
more transparency in features which the classifiers use to identify particular classes.

\begin{figure}[!hbt]
    \centering
    \includegraphics[width=.65\textwidth]{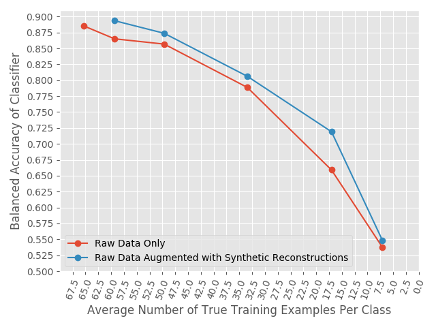}
    \caption{\textbf{Comparison of classifier performance on raw data only and raw data augmented with synthetic examples from
    the DAPPER GAN system.} \emph{As we decrease the number of examples per class (x-axis), we find that, while the
    performance of the classifier with guided data augmentation consistently outperforms the classifier trained on raw
    data, achieving as muchas a 6\% increase over the classifier trained on raw data only.}}
    \label{fig:classifier_perf}
\end{figure}

We showed the DAPPER GAN prototype directly enhances classification performance on a sparse unseen dataset containing 10
fine-grained vehicle makes and models (e.g., BMW X3, BMW X5, Acura TL, Acura TSX). The prototype projects real training
images into a shared latent space and randomly perturbs this projection before reconstructing the input to synthetically
augment training data quantity and increase data diversity. While we expect semantic manipulation (e.g., 3D rotation)
will have greater impact, we did achieve improved classification accuracy using random latent space perturbations,
requiring up to 30\% less real training data to achieve comparable or better results than using raw data alone. We
illustrate the results of our 5-fold cross validation experiment using a ResNeXt classifier pretrained on ImageNet in
Figure
\ref{fig:classifier_perf}. 

\begin{figure}[!hbt]
    \centering
    \includegraphics[width=.5\textwidth]{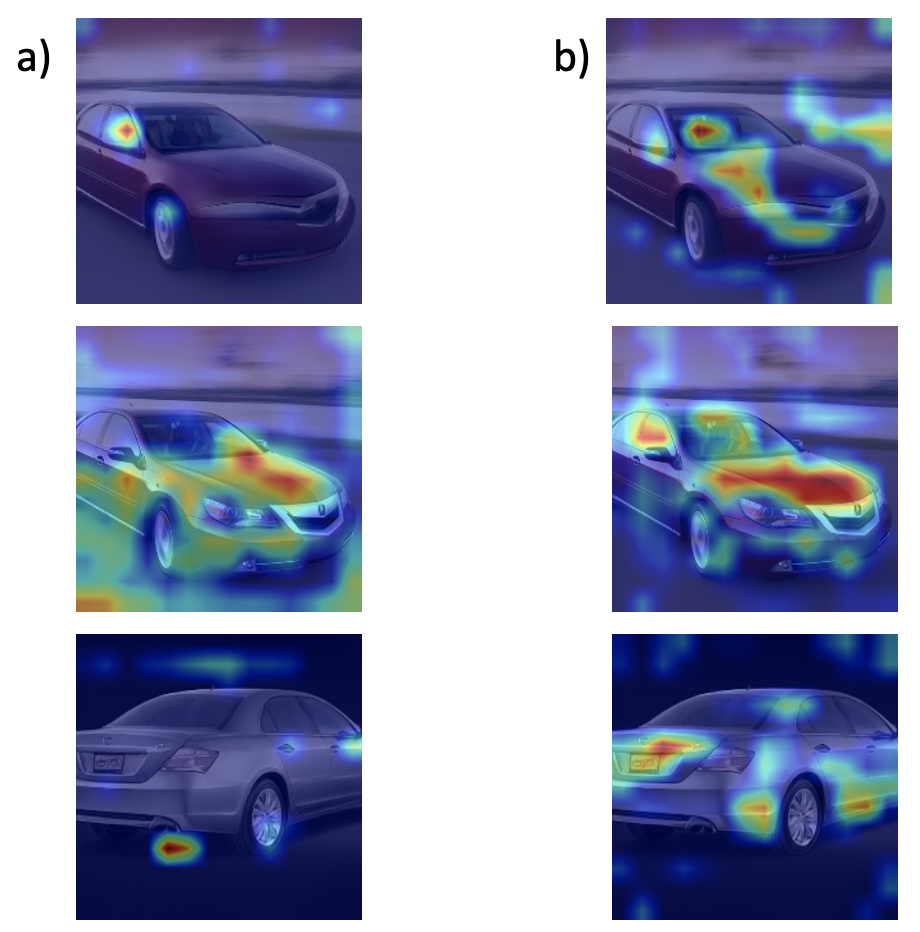}
    \caption{\textbf{Grad-CAM saliency maps.} \emph{On the left, is a Grad-CAM attention map for our ResNext vehicle
            classifier for real data only; the right shows a Grad-CAM attention map for our ResNext classifier with real
            plus synthetic data.  The red regions correspond to areas of the input image the classifier found more
            important in making a classification.  In the right image (real plus synthetic), it is clear the network is
            attending to more meaningful features that on the right (e.g., the left model is attending to the ground).}}
    \label{fig:xai}
\end{figure}

We furthermore showed that Grad-CAM based attention maps homed in on semantically meaningful image features after
synthetic augmentation, and also qualitatively improved which features were attended to.~\cite{selvaraju_grad-cam_2020}
Example images are shown in Figure \ref{fig:xai}.

\FloatBarrier

\section{Discussion and Future Works}
\label{sec:discussion}

In this paper, we demonstrated the feasibility of transferring information from one dataset to another through the use
of StyleGAN2-based dataset augmentation. Furthermore, we showed that this transfer of information led to an increase in
downstream ML tool performance, allowing the use of less training data to achieve similar evaluation performance on the
target dataset. In particular, we showed the DAPPER GAN prototype directly enhances classification performance on a
sparse unseen dataset containing 10 fine-grained vehicle makes and models (e.g., BMW X3, BMW X5, Acura TL, Acura TSX).
The prototype projects real training images into a shared latent space and randomly perturbs this projection before
reconstructing the input to synthetically augment training data quantity and increase data diversity. While we expect
semantic manipulation (e.g., 3D rotation) will have greater impact, we did achieve improved classification accuracy
using this method, requiring up to 30\% less real training data to achieve comparable or better results than using raw
data alone. We furthermore developed a semantic rotation technique which can effectively rotate vehicles to provide
novel data views. Future work includes testing this rotation technique on applicable datasets and developing further
robustness of the rotation technique.

\section{Acknowledgements}
\label{sec:acknowledge}
This work was performed under USA-ACC-APG contract number W56KGU-19-C-0031 with the U.S. Army Combat Capabilities
Development Command C5ISR Center. The authors thank Mr. Flloyd Cuttino, and Mr. Omar Solaiman for
their support and direction on this project.

\bibliography{dapper_gan}{}
\bibliographystyle{unsrt}
\end{document}